# Human-Aligned Evaluation of a Pixel-wise DNN Color Constancy Model

Hamed Heidari-Gorji, Raquel Gil Rodriguez, Karl R. Gegenfurtner
Psychology Department, Giessen University, Alter-Steinbacher-Weg 38, Giessen, 35394, Germany.

Correspondence to: gegenfurtner@uni-giessen.de, hamed.h@live.com

## Abstract

We previously investigated color constancy in photorealistic virtual reality (VR) and developed a Deep Neural Network (DNN) that predicts reflectance from rendered images. Here, we combine both approaches to compare and study a model and human performance with respect to established color constancy mechanisms: local surround, maximum flux and spatial mean. Rather than evaluating the model against physical ground truth, model performance was assessed using the same achromatic object selection task employed in the human experiments. The model, a ResNet based U-Net from our previous work, was pre-trained on rendered images to predict surface reflectance. We then applied transfer learning, fine-tuning only the network's decoder on images from the baseline VR condition. To parallel the human experiment, the model's output was used to perform the same achromatic object selection task across all conditions. Results show a strong correspondence between the model and human behavior. Both achieved high constancy under baseline conditions and showed similar, condition-dependent performance declines when the local surround or spatial mean color cues were removed.

## 1. Introduction

Color constancy refers to the ability to perceive stable surface colors despite changes in the spectral composition of the illumination. Although the light reaching the eye varies substantially with illumination, human observers nevertheless maintain relatively consistent representations of object color across a wide range of viewing conditions. This stability is central to object recognition and memory, and has motivated decades of research into the perceptual and computational mechanisms underlying color constancy (Gegenfurtner and Kiper, 2003; Smithson, 2005; Ebner, 2007; Brainard and Maloney, 2011; Foster, 2011; Witzel and Gegenfurtner, 2018; Hurlbert, 2019; Maloney and Knoblauch, 2020).

Researchers have identified three primary mechanisms contributing to color constancy. The first, the local surround mechanism, posits that variations in color across local edges tend to be relatively constant under varying illuminant changes. This was exemplified by the finding that the color appearance of a central disk depends upon the luminance ratio between center and surrounding ring, irrespective of their absolute luminance levels (Wallach, 1948). Subsequent research has generalized this to color (Valberg and Lange-Malecki, 1990; Foster and Nascimento, 1994). The second mechanism, maximum flux ("bright is white"), states that the visual system compensates for the brightest area of a scene and assumes it to be white (Land and McCann, 1971; Cataliotti and Gilchrist, 1995; Gilchrist and Radonjić, 2010). A third mechanism is the spatial mean ("gray world") and adaptation to the global mean chromaticity of a scene on the assumption that its average reflectance is typically achromatic (Buchsbaum, 1980; Brainard and Wandell, 1986). Importantly, these cues are not used in isolation. Human color constancy reflects a context-dependent integration of multiple, partially redundant sources of information whose reliability depends on scene structure, illumination, and task demands (Kraft and Brainard, 1999; Radonjić and Brainard, 2016).

While traditional computer vision approaches to color constancy used similar cues (Land and McCann, 1971; Buchsbaum, 1980; see Ebner, 2007 for review), most recent developments in color constancy modeling have shifted toward data-driven approaches, particularly convolutional neural networks (Bianco et al., 2015, 2017; Lou et al., 2015; Hu et al., 2017; Afifi et al., 2021; Domislović et al., 2022; Buzzelli et al., 2023; Wang et al., 2023). These models were trained either to estimate the scene illuminant directly or to predict surface reflectance from image data, often outperforming traditional heuristic methods in complex, unconstrained environments. Work from several groups has established that CNN-based approaches can exploit spatial context, object boundaries, and image statistics in ways that are difficult to formalize analytically, leading to substantial performance gains over classical algorithms (see Bianco et al., 2015). More recently, the landscape of computational color constancy has further expanded to include biologically-inspired learning-free methods (Ulucan et al., 2022, 2024), spatially-varying illuminant estimation for mixed lighting (Entok et al., 2024; Kim et al., 2024), and spectral-informed approaches (Erba et al., 2024). Extensions to fully convolutional encoder–decoder architectures further enable pixel-wise predictions of surface color, bridging the gap between global illuminant estimation and object-level constancy (Flachot et al., 2022; Heidari-Gorji and Gegenfurtner, 2023).

Despite these advances, an important conceptual distinction remains between the goals of computational models and those of human vision research. Behavioral studies of color constancy have traditionally aimed to characterize how well human observers maintain stable color appearance, and how performance depends on the availability of specific cues. In contrast, most computer vision models are optimized to recover physical ground truth, such as the scene illuminant or surface reflectance, and are evaluated accordingly. While these objectives often overlap, they are not equivalent. In particular, it is now well established that human observers can exhibit good color constancy while being relatively poor at explicitly estimating the illuminant(Rutherford and Brainard, 2002; Smithson, 2005; Reeves et al., 2008; Granzier et al., 2009; Radonjić et al., 2015). This dissociation implies that accurate illuminant estimation is neither necessary nor sufficient for human-like color constancy. Consequently, while many computational approaches frame color constancy as an illuminant estimation problem, the present work adopts a different perspective. Rather than evaluating models against physical ground truth, we assess whether a model reproduces the structure of human behavior across manipulations of cue availability and scene context, using human performance as the primary reference.

We have previously investigated the contributions of local surround, maximum flux, and spatial mean mechanisms within an immersive virtual reality (VR) environment (Rodríguez et al., 2024). In these experiments, five illuminants were used, three along the daylight locus (neutral, blue, yellow) and two orthogonal to it (red, green), and specific scene manipulations were employed to selectively disrupt individual cues. Local surround information was eliminated by placing the target object on a surface with fixed chromaticity, independent of illumination. The maximum flux cue was silenced by introducing a bright object with constant neutral chromaticity, and the spatial mean was controlled either by adding objects to the scene or by modifying the reflectances of existing objects. Crucially, the VR setup provides access to the same rendered images seen by the observers, allowing direct evaluation of an image-based, pixel-wise color constancy model under identical conditions (Heidari-Gorji and Gegenfurtner, 2023).

# 2. Methodology

This section describes the methodology employed to investigate color constancy in VR and to compare human perceptual responses with those predicted by computational models. Section 2.1 presents the psychophysical experiments conducted with human observers in immersive VR environments, outlining the experimental design, stimuli, and analysis of perceptual data. Section 2.2 details the proposed color constancy model, including its architecture, training strategy, perceptual loss formulation, and adaptation to replicate the psychophysical task. Section 2.3 introduces the experiments performed to evaluate the model under controlled conditions and to compare its behavior with human data. Finally, Section 2.4 describes the benchmark color constancy algorithms used for comparison, ranging from classical statistical approaches to modern deep learning methods.

## 2.1 Color Constancy Psychophysical Experiments in Virtual Reality

In our previous work, we investigated color constancy in virtual reality, exploring several well-established mechanisms such as local surround, maximum flux, and spatial mean (Rodríguez et al., 2024). We designed a selection task experiment using five colored illuminants: three aligned with the daylight locus (blue, neutral, and yellow) and two along the orthogonal axis (red and green) (Aston et al., 2019). Participants were initially shown an achromatic target object under a neutral illuminant and instructed to memorize its appearance. They were then exposed to one of the five test illuminants (presented in random order) and given one minute for adaptation. After this period, the target object, along with four competitors, appeared in different positions within the scene. Participants were asked to select the object they perceived as matching the original target. Each participant completed 30 trials per adaptation condition, after which the process restarted with a new presentation of the achromatic target under the neutral illuminant.

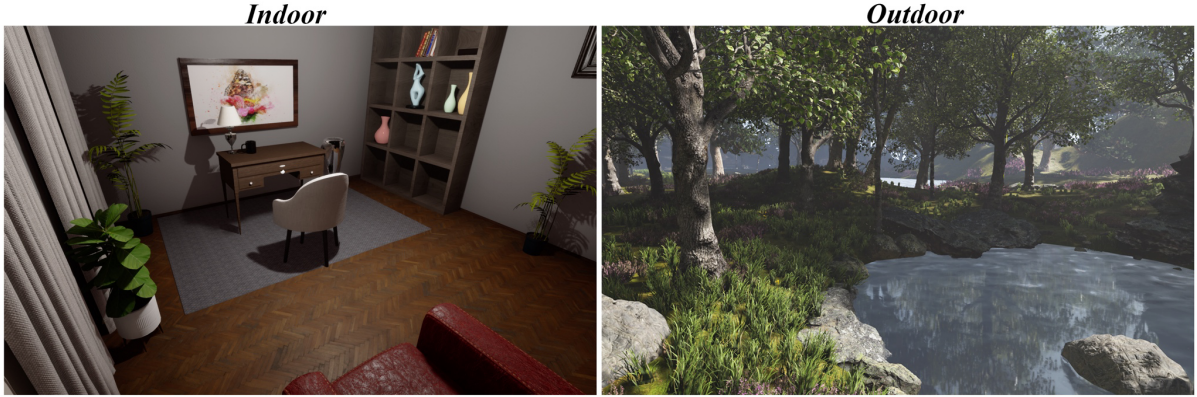

*Figure 1. Experimental scenes rendered in virtual reality (Rodríguez et al., 2024). The indoor office scene (left) features artificial illumination, while the outdoor landscape (right) simulates natural daylight conditions with directional sun and skylight.*

The experiments were conducted in two distinct virtual environments: an indoor office-like scene and an outdoor natural landscape featuring cliffs, trees, grass, and a lake (see Figure1). The indoor scene was illuminated by a single point light source positioned at the center of the room. In contrast, the outdoor scene included both a directional light simulating sunlight and an additional skylight component to create a more realistic outdoor lighting environment.

The illumination colors, as well as the colors of the target and competitor objects, were carefully characterized following the methodology described in (Gil Rodríguez et al., 2022). Competitor colors were defined under each specific illuminant as follows:

● The reflectance match (R) shares the same surface reflectance as the target, and it is considered perfect color constancy.

● The tristimulus match (T) was defined such that its reflected light under the current illuminant matched the reflected light of the target under the neutral illuminant, considered zero color constancy.

● Between these two (S1, S2), we introduced two intermediate samples, equally spaced in the CIELAB color space.

● Finally, an over constancy (O) sample was defined along the same color axis but in the direction opposite to the reflectance match.

The collected data were analyzed using the adapted Maximum Likelihood Difference Scaling (MLDS) method from (Radonjić et al., 2015), which estimates the perceptual positions of the competitors and the participant's inferred match along a one-dimensional perceptual scale. The inferred match, **Match** from now on, can then be mapped into CIELAB space by preserving its relative position with respect to the competitors. Based on these inferred matches, we compute in CIELAB color space the following Color Constancy Index (CCI) in what we called the *competitor's color space*:

$$CCI = \frac{|T - Match|}{T - R}, \quad (1)$$

where T corresponds to the tristimulus value, R to the reflectance, and Match to the participants' inferred match. We report the percentage CCI by multiplying the above equation by 100. This means that R represents 100% color constancy, while T represents 0% color constancy.

A total of 20 naive participants conducted the experiments, with 10 for the indoor setting and 10 for the outdoor setting.

## 2.2 Color Constancy Model

A model capable of labeling every point in the input image is required for this project. This purpose can be achieved by using segmentation models. The U-Net models are widely used in semantic or instance segmentation. The input to this model is an image with one or multiple channels of color coordinates. The output shows which pixel or part of the input belongs to which object. A layer in the output represents each object. Each pixel in the output corresponds to a portion or pixel of the input image. This model comprises an encoder and a decoder. Various DNN models can be used for these two parts. As an encoder, a ResNet with 50 layers was used in our model.

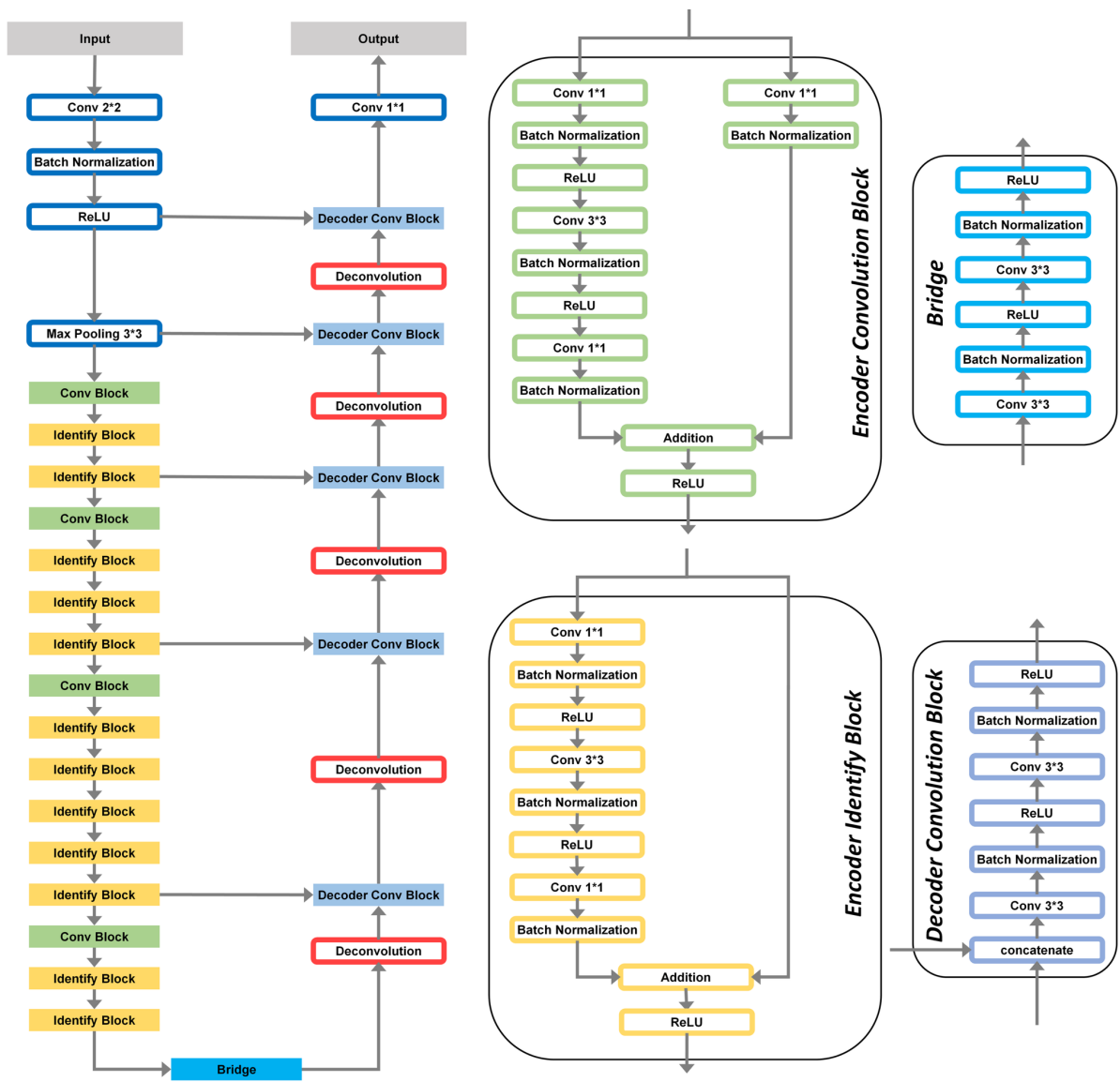

*Figure 2. Architecture of the color constancy model proposed by (Heidari-Gorji and Gegenfurtner, 2023). The model uses a ResNet-50 backbone as an encoder to process the RGB input and a decoder*

*to reconstruct the surface reflectance in CIELAB color space. The right panels detail the internal structure of the four key components: the Encoder Convolution Block and Encoder Identify Block (used in the downsampling path), the Bridge (connecting the encoder to the decoder), and the Decoder Identify Block (used in the upsampling path to predict CIELAB reflectance).*

The model is designed to recognize the inherent reflectance color of an input image, regardless of the lighting conditions. The input to this model is an image in the RGB color space, which consists of three layers. The model's output is an image in the CIE-Lab color space.

## 2.2.1 Perceptual Balanced Color Loss (PBCLoss)

Color-constancy networks must learn to predict intrinsic reflectance color in the perceptually uniform CIELAB space; however, the training distribution is heavily skewed. In natural images, most pixels lie close to the achromatic axis, while highly saturated hues are rare. When the loss is a plain mean-squared error (MSE) on each channel, the model can minimize risk simply by shrinking all a and b values, producing desaturated outputs that nevertheless achieve a low numeric error. To address this imbalance, we formulate Perceptual Balanced Color Loss (PBCLoss), a composite objective that measures perceptual fidelity with CIEDE2000(Sharma et al., 2005) while explicitly amplifying the penalty on saturated chroma, while still anchoring luminance.

For every pixel, we compute the CIEDE2000, and we add a chroma-weighted squared error on the chromatic channels, where the weight is:

$$\omega_i = 1 + \beta \left(\frac{C_i^{gt}}{128}\right)^\gamma, \quad (2)$$

$$C_i^{gt} = \sqrt{\left(a_i^{gt}\right)^2 + \left(b_i^{gt}\right)^2}, \quad (3)$$

Boosts the contribution of rare, high-chroma pixels (Milidonis et al., 2025). Combining these terms with a lightness constraint yields.

$$PBCLoss = \lambda_1 \Delta E_{00} + \lambda_2 \omega_i \left[\left(a_i^p - a_i^{gt}\right)^2 + \left(b_i^p - b_i^{gt}\right)^2\right] + \lambda_2 \left(L_i^p - L_i^{gt}\right)^2, \quad (4)$$

Where the over-bar averages over all pixels in the mini-batch. In our experiments, we set $\lambda_1 = 1$, $\lambda_2 = 0.5, \lambda_3 = 0.2$ and $\beta = 2$, $\gamma = 2$, values chosen to balance vivid color reproduction against photometric stability.

## 2.2.2 Application to the psychophysical experiment

In the human experiment, observers performed a selection task in which they had to identify the reference (achromatic) lizard from a group of five competitors, under various illuminants and across multiple trials (see Section 2.1 for details).

To replicate the experiment conducted with human observers, we trained our color constancy model using the baseline condition of the two scenes. Specifically, we froze the encoder and trained only the decoder.

Since the scenes were rendered in Unreal Engine, we had access to both the final rendered images and their corresponding intrinsic reflectance values, needed inputs for our model. Each scene was rendered under five illuminants. For every illuminant we generated five trials, and within each trial the scene was rendered from five distinct viewpoints.

To adopt this task for the model, we captured images of each color competitor lizard under each illuminant, across all the spatial locations where lizards had appeared in the scenes. This resulted in a set of images for each color competitor, with the same lizard color shown in different positions. For a given illuminant, the full set of images, containing all color competitors across all locations, served as input to the model.

By masking each lizard from the output of the model, and averaging the reflectance values across all its appearances, under one specific illuminant, we obtained a single reflectance value per competitor. As the model outputs reflectance is in the CIELAB color space, this process yields one LAB value for each competitor, representing the model's "perceived" color for that lizard.

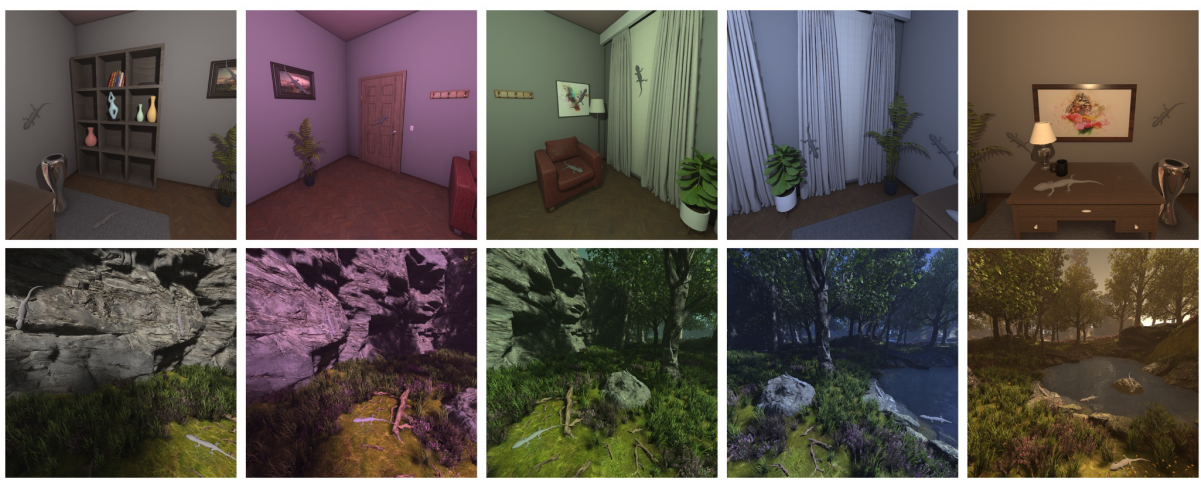

*Figure 3. The Baseline Image Dataset for Decoder Fine-Tuning. The images presented are inputs for the transfer learning stage. Each column corresponds to a unique lighting condition applied to a pair of scenes: an indoor environment (top) and an outdoor environment (bottom). These images represent the control baseline, characterized by the absence of any disruption to the color constancy cues, and were used to fine-tune the model's decoder.*

## 2.2.3 From model output to color constancy index: the model color constancy index

Now that we have an LAB color for each competitor from the model's output, we need to define the model's Match. Let us consider the a-b chromatic plane with the original values of the competitors' reflectance, together with the values of the competitors from the model's output, as described above. See left side of Figure 4, where the ringed yellow circles represent the output values of the model, and the non-ringed the original ones. Remember that R refers to reflectance (100%) and T to tristimulus (0%).

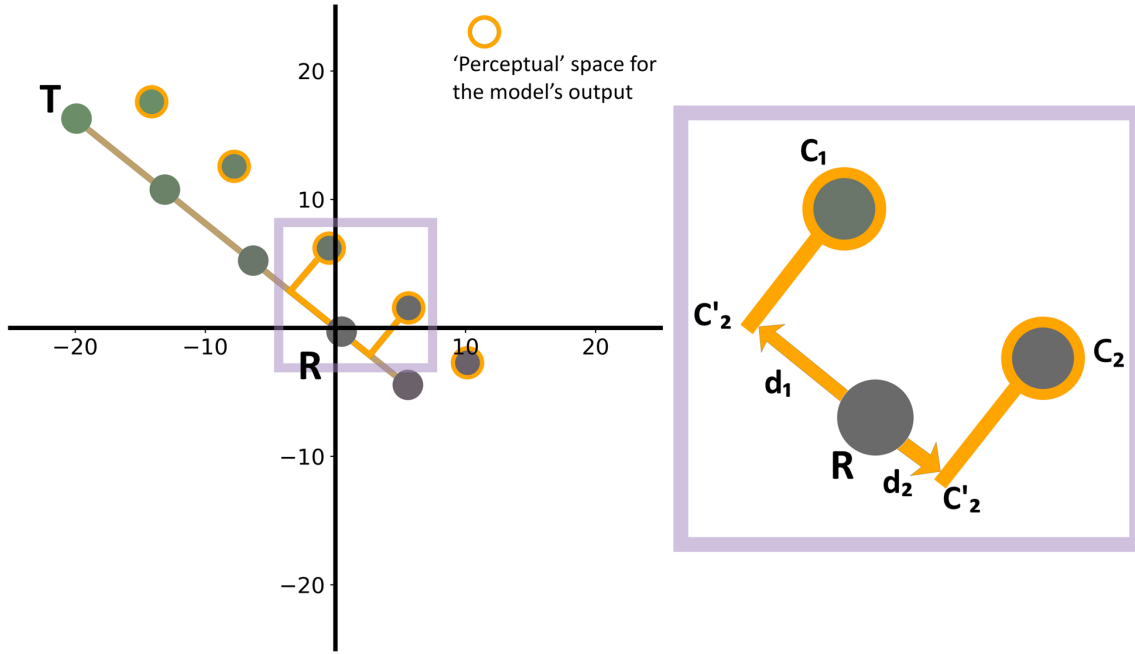

*Figure 4. Method for deriving the model's inferred match in CIELAB color space. Left: The a\*–b\* chromaticity plane displaying the set of experimental competitors (solid gray spheres) and the corresponding model predictions (yellow ringed spheres). The axis spans from the Tristimulus match (T, 0% constancy) to the Target Reflectance (R, 100% constancy). Right: A zoomed-in view illustrating the projection and interpolation logic. The closest model outputs are projected onto the competitor axis (points $C'_1$ and $C'_2$), and their distances ($d_1$ and $d_2$) from the Reflectance point (R) are used to calculate the final match.*

To determine the model's Match, its estimated perception of the achromatic lizard. We followed these steps:

**1. Projection**: For each competitor output we projected it onto the line connecting the reflectance point R and T.

$$Proj_{RT}(P) = R + \frac{(T-R).(P-R)}{\|T-R\|^2}, \quad (5)$$

**2. Distance Calculation**: We computed the distance from each projected to the reflectance point R, which corresponds to the achromatic color.

**3. Selection of Nearest Competitors**: We identified the two competitors whose projected outputs were closest to R.

**4. Interpolation**: Using the relative distances to R, we interpolated between the original competitors of the corresponding $C_m$ to obtain the model's Match.

**5. Color constancy index**: the CCI is computed by using the computed Match in Equation 1.

## 2.3 Experiments

We conducted two experiments which directly compare the output of the color constancy model with human observer data.

## 2.3.1 Experiment 1: Color constancy mechanisms

Three color constancy mechanisms were investigated in (Rodríguez et al., 2024) **local surround**, 2) **maximum flux**, and 3) **spatial mean**. The **local surround** mechanism refers to chromatic adaptation influenced by the immediate background of the target. To suppress this, the surround color was held constant across changes in illumination. The **maximum flux** mechanism, also known as the "white patch", assumes the visual system adapts to the brightest region in the scene, interpreting it as the illuminant. Suppression was achieved by maintaining a constant achromatic region across illuminants. The **spatial mean** mechanism, also referred to as the "gray world" assumption, is based on adaptation to the average color of the entire scene. This was suppressed in two ways: (1) by adding new objects with colors opposing the illuminant direction in a color-opponent space, and (2) by modifying the reflectance of existing objects to shift the overall scene mean in the opposite direction of the illuminant. See Figure 5.

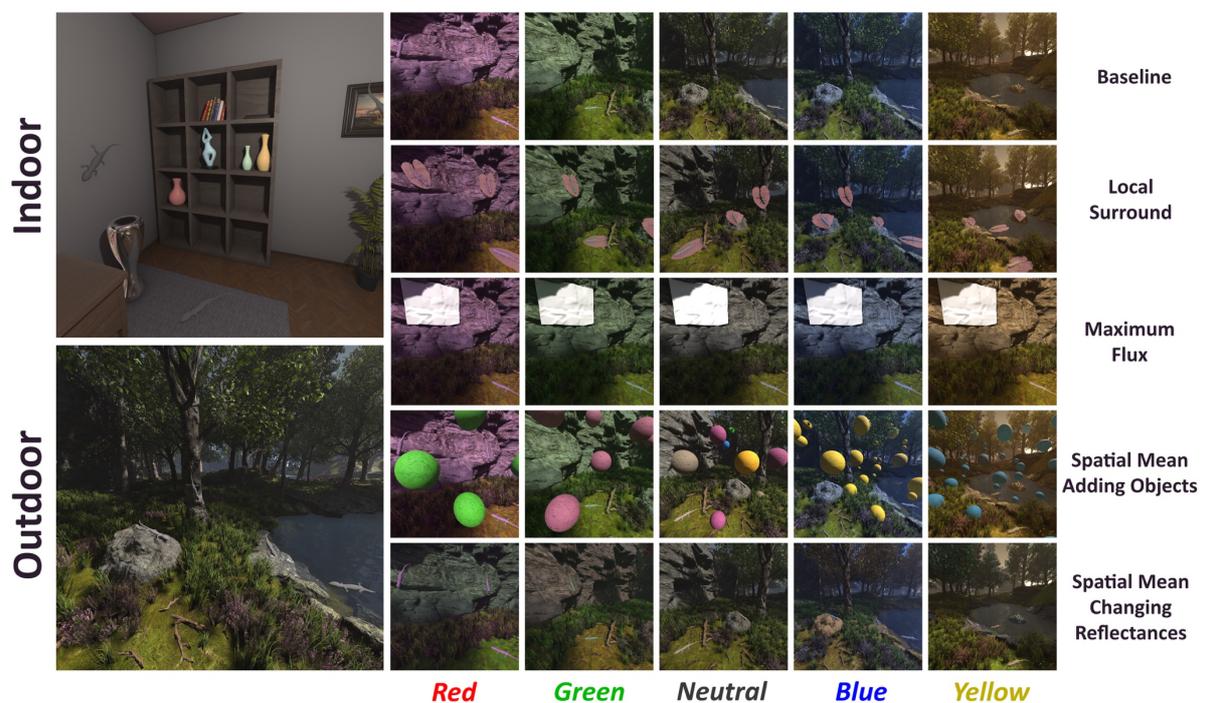

*Figure 5. Color constancy mechanism design in VR. On the left side, there are both indoor and outdoor scenes under neutral illumination. The right side showcases all five illuminants (red, green, neutral, blue, and yellow) arranged in columns. The rows illustrate each mechanism: baseline, local surround (target placed on constant pink leaf), maximum flux (adding a constant white patch), spatial mean with added objects (floating colored spheres), and spatial mean with changing reflectances (changing colors of the environment), all depicted in the outdoor scene.*

Observers performed all the different mechanisms and designs together with a baseline experiment, where all cues were present in the scene. The experiments were done in indoor and outdoor scenes, under five different illuminations.

For the model's data, under each condition, we defined a set of images with all the lizard competitor colors in all possible locations, as shown in Figure X. Then, as described in Sections 2.2.2 and 2.2.3, we obtained for each condition and illuminant the model's Match, that can be compared directly with the participants data.

## 2.3.2 Experiment 2: Interaction between local surround color and illumination

In the second experiment, we investigated the Local Surround mechanism. This time, we used four different colored surrounds: *khaki*, *rose*, *purple*, and *teal*. Figure 6 illustrates the illuminants, target circles, and the four surround colors (the leaves), all plotted in the *a\*-b\** plane of the CIELAB color space. We defined them along the diagonals of the illuminants, same as in our previous Experiment 1 (Aston et al., 2019), in CIELAB color space.

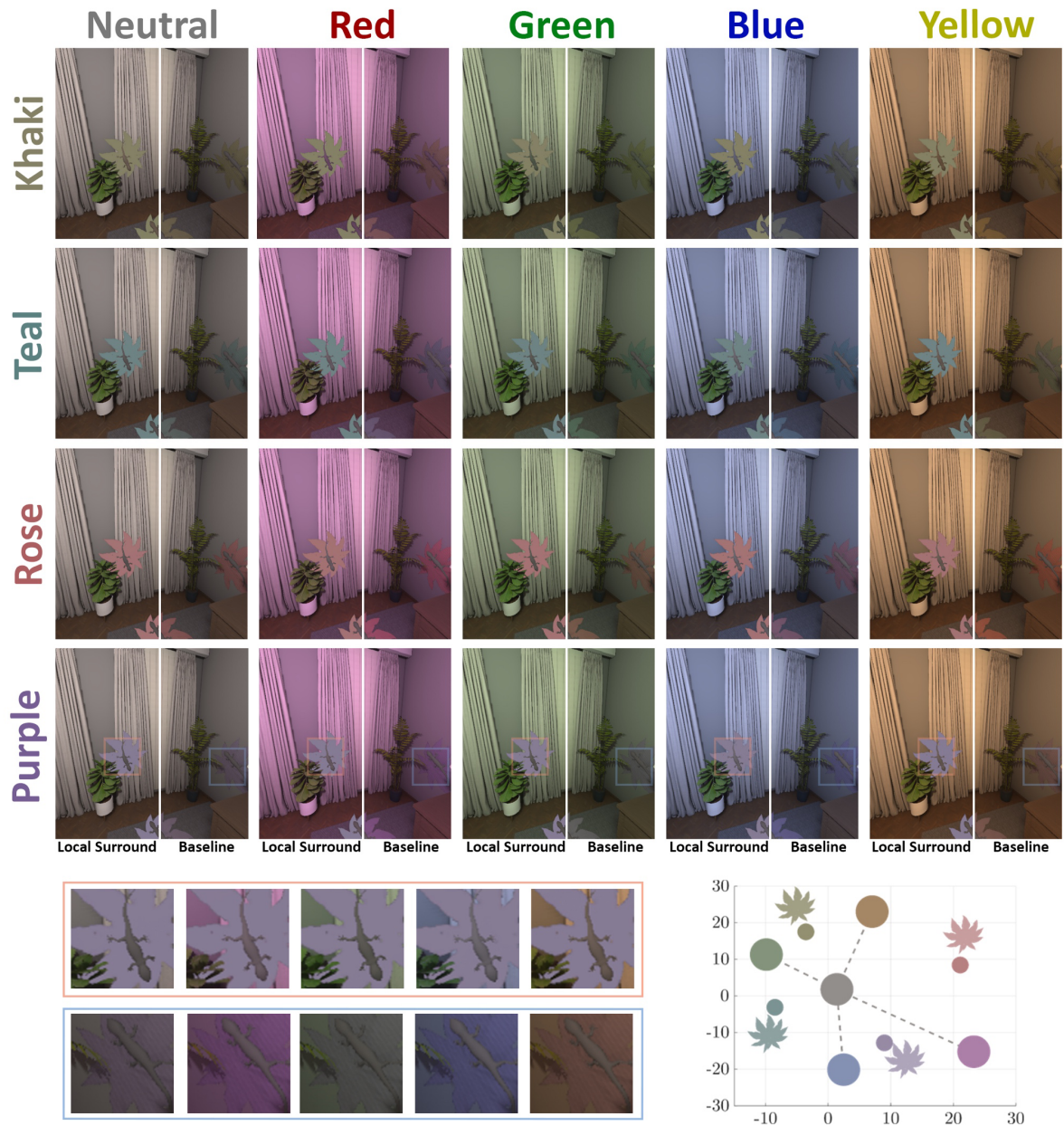

*Figure 6. Experimental stimuli and colorimetric specification for the local surround experiment. The main grid of images displays the rendered scenes, organized by four local surround leaf colors (rows: Khaki, Teal, Rose, and Purple) and five illumination conditions (columns: Neutral, Red, Green, Blue, and Yellow). For each condition, the image on the left demonstrates our "Local Surround" mechanism, where the leaf color is held constant across illuminations, while the image on the right serves as the "Baseline" where the leaves are naturally affected by the illuminant. The*

*plot at the bottom-right specifies the chromaticities of the key components in the CIELAB space, where circles represent the illuminants and leaf shapes represent the leaf colors; the axes correspond to the a\* and b\* dimensions. To facilitate detailed inspection, the bottom-left panel shows zoomed-in views of the regions highlighted by orange and blue frames within the Purple leaf colors row, allowing for a close examination of the visual differences between the "Local Surround" and "Baseline" images.*

In this experiment, 16 participants completed the **Local Surround** task for each of the four surround colors, along with a corresponding baseline condition specific to each color. In these scenes, all lizards were placed on leaves of the given surround color, unlike the baseline of Experiment 1, where no leaves were placed under the lizards. Now the leaf's surface reflectance interacted with the illuminant, causing the leaves to reflect different colors under different lighting conditions. In the experimental condition, the mechanism was suppressed in the same way as in Experiment, by keeping the reflected color of the leaf constant, regardless of changes in illumination. The experiments were done in indoor and outdoor scenes, under five different illuminations.

For the model's data, under each condition, we defined a set of images with all the lizard competitor colors in all possible locations, as shown in Figure 6. Then, as described in Sections 2.2.2 and 2.2.3, we obtained for each condition and illuminant the model's Match, which can be compared directly with the participants' data.

## 2.4 Comparison with color constancy models

To evaluate the performance of our proposed model, we compare it against six prominent color constancy algorithms, ranging from classical statistical methods to a recent deep learning-based approach: 1) gray world, 2) white patch, 3) shades of gray, 4) gray edge, 5) weighted gray edge and 6) a mixed illuminant white balance model. All of these methods provide as an output a white balance version of the input image under what they consider as a neutral illumination. For the statistical methods (1-5), we utilized the implementations provided by Gijsenij et al. (Gijsenij et al., 2011).

1. The *Gray World* algorithm is a foundational method based on the assumption that the average reflectance of a scene is achromatic (gray). The illuminant's color is estimated by calculating the deviation from this gray average and correcting accordingly (Buchsbaum, 1980).

2. The *White Patch* (or Max-RGB) algorithm operates on the principle that the brightest patch in a scene is a perfect white reflector. Therefore, the maximum sensor response in each channel (R, G, B) is assumed to directly correspond to the color of the light source (Land, 1977).

3. The *Shades of Gray* method provides a generalized framework that encompasses both Gray World and White Patch. It utilizes the Minkowski p-norm to estimate the illuminant, where different values of 'p' can shift the algorithm's behavior between averaging and maximum-value selection (Finlayson and Trezzi, 2004).

4. The *Gray Edge* algorithm shifts the focus from pixel intensities to their derivatives. It assumes that the average of the color differences at the edges within an image is neutral. This makes it more robust to large, uniformly colored areas that can bias other statistical methods (Van De Weijer and Gevers, 2005; Van De Weijer et al., 2007).

5. A refinement of this concept is the *Weighted Gray Edge* model. It improves upon the Gray Edge method by assigning more weight to more prominent edges (i.e., those with a larger magnitude), thereby reducing the influence of noise and minor textures in the illuminant estimation (Gijsenij and Gevers, 2007).

6. Finally, we include the *Mixed-Illuminant White Balance* model, a convolutional neural network created for complex scenes with multiple light sources. This model learns to correct spatially varying illumination, representing a state-of-the-art benchmark for complex lighting (Afifi et al., 2022).

# 3. Results

We begin with a structured overview of the main findings to orient the reader before presenting the detailed statistical analyses. Because our goal is not simply to report model performance, but to evaluate whether the model reproduces the *structure* of human color constancy across controlled cue manipulations, we first summarize the core patterns observed across experiments.

Across both experiments, human observers and the model showed high levels of color constancy under baseline conditions, in both indoor and outdoor scenes. Systematic removal of individual constancy cues revealed condition-specific degradations in performance. In Experiment 1, disrupting spatial mean information produced the largest declines in constancy, followed by local surround manipulations, whereas maximum flux had comparatively modest effects. In Experiment 2, suppressing the local surround mechanism led to direction-dependent impairments: performance declined primarily when the illuminant was chromatically opposing the surround, while neighboring illuminants produced little change.

Crucially, the model reproduced not only the overall magnitude of human constancy under baseline conditions, but also the relative pattern of performance changes across mechanisms, illuminants, and scene contexts. This structural similarity distinguishes it from classical global-statistics algorithms, which often achieve comparable average constancy yet fail to capture the selective, cue-dependent degradations observed in human behavior. The following sections provide a detailed analysis of these effects.

The boxplots in all figures display the CCI performance of participants (dots) and the model (star), along with the overall mean (dashed line) and median (solid line), bounded by the 25th and 75th percentiles.

## 3.1 Experiment 1: Color constancy mechanisms

We aim to analyze how the suppression of each mechanism affects CCI performance. To achieve this, we compute the CCI difference, ΔCCI, which is defined as the difference between CCI performance under a specific mechanism and CCI performance under the Baseline condition, where all cues are available. We will use a Generalized Linear Mixed-Effects (GLME) model, treating ΔCCI as the response variable and the different mechanisms as fixed effects, while considering participants as random effects. For visualization and comparative purposes, we treated the model as an additional observer.

*Table 1. Statistical ΔCCI summary of color constancy mechanisms in indoor and outdoor scenes, showing human observer performance alongside the model's responses treated as an additional observer.*

| Mechanism | Scene | Estimated Mean | 95% CI (Lower, Upper) | p-value |
|---|---|---|---|---|
| Spatial Mean – Changing Reflectances | Indoor | −75.78 | [−82.47, −69.08] | < 0.0001 |
| Spatial Mean – Adding Objects | Indoor | −23.03 | [−30.41, −15.64] | < 0.0001 |
| Maximum Flux | Indoor | −7.29 | [−13.71, −0.88] | 0.026 |
| Local Surround | Indoor | −23.19 | [−29.60, −16.77] | < 0.0001 |
| Spatial Mean – Changing Reflectances | Outdoor | −68.30 | [−78.66, −57.94] | < 0.0001 |
| Spatial Mean – Adding Objects | Outdoor | −16.05 | [−26.08, −6.02] | 0.002 |
| Maximum Flux | Outdoor | −2.09 | [−11.83, 7.64] | 0.672 |
| Local Surround | Outdoor | −14.57 | [−24.31, −4.84] | 0.004 |

Results in Table 1 collectively demonstrate that spatial and local surround mechanisms strongly influence perceived color constancy in both environments, with the effect of maximum flux being scene-dependent.

Figure 7 shows the CCI under each of the mechanisms, together with the baseline in the indoor (left) and outdoor (right) environments. Each dot represents a participant, and the star the model. Figure 8 presents each mechanism separately, shown by colored illuminant, for indoor (first row) and for outdoor (second row).
A limitation of our color constancy model is that, in the spatial mean adding objects condition, closely clustered competitor estimates can produce highly variable CCI values, sometimes yielding misleadingly high or low scores despite nearly indistinguishable perceptual outputs, as shown in Figure 8, both indoor and outdoor scenes.

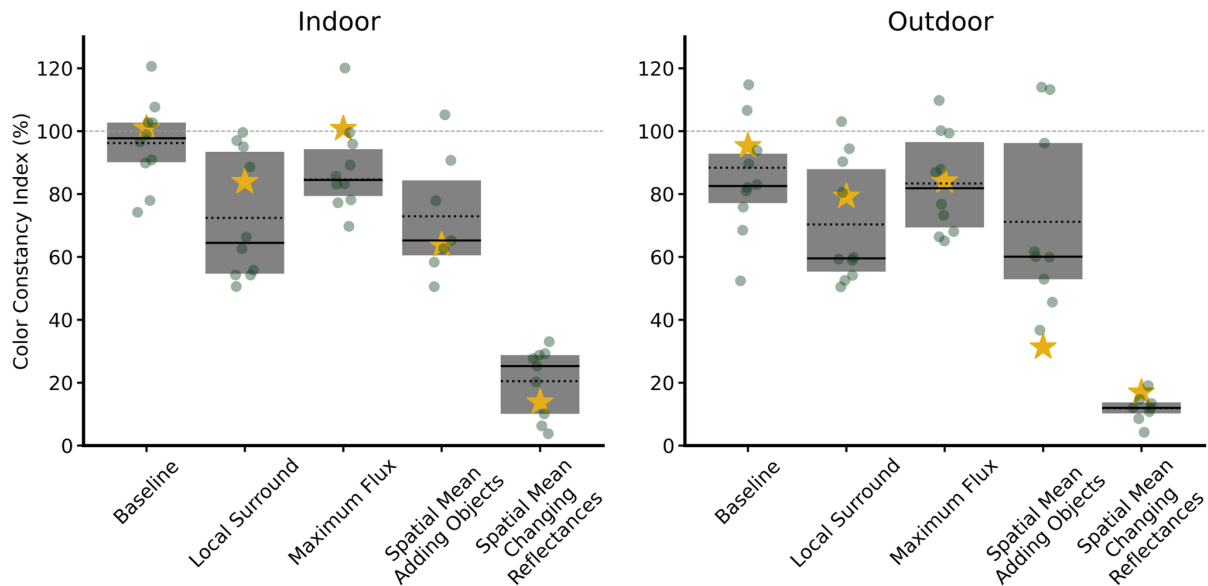

*Figure 7. Color constancy index (CCI) percentage in all conditions. For each condition, results are shown for indoor (left) and outdoor (right) environments. Each gray dot represents the mean performance of one subject across all illuminations, and the yellow star indicates the mean of the model across all illuminations. The boxplots summarize the human data: the solid line indicates the median, and the dotted line indicates the mean. The horizontal dashed line at 100% represents perfect constancy.*

We also calculated the intra- and inter-participant variability for both indoor and outdoor scenarios, as these involved different groups of participants. To assess inter-participant variability, we computed the Pearson correlation coefficient based on the CCI values for each participant under each mechanism, then averaged the Pearson coefficients across all participants. For the intra-participant variability, we estimated the coefficient of variation, which is the standard deviation divided by the mean, for each participant across all mechanism conditions.

In the indoor scene, the inter-participant variability shows a very high correlation of over 0.70, with one exception that exhibits a negative correlation. In the outdoor scene, we found that the inter-participant variability is above 0.50 for all participants, except for one individual who also has a negative correlation with the rest of the group. In both scenarios, the model showed consistent results, with correlation coefficients of 0.75 for the indoor setting and 0.51 for the outdoor setting. Notably, we observed higher correlations in the indoor performance.

The intra-participant variability indicated that there is more variability in the outdoor scene, with values ranging from 0.136 to 0.5517 and a mean value of 0.326. In contrast, the indoor scene ranged from 0.099 to 0.933, with a lower mean value of 0.260.

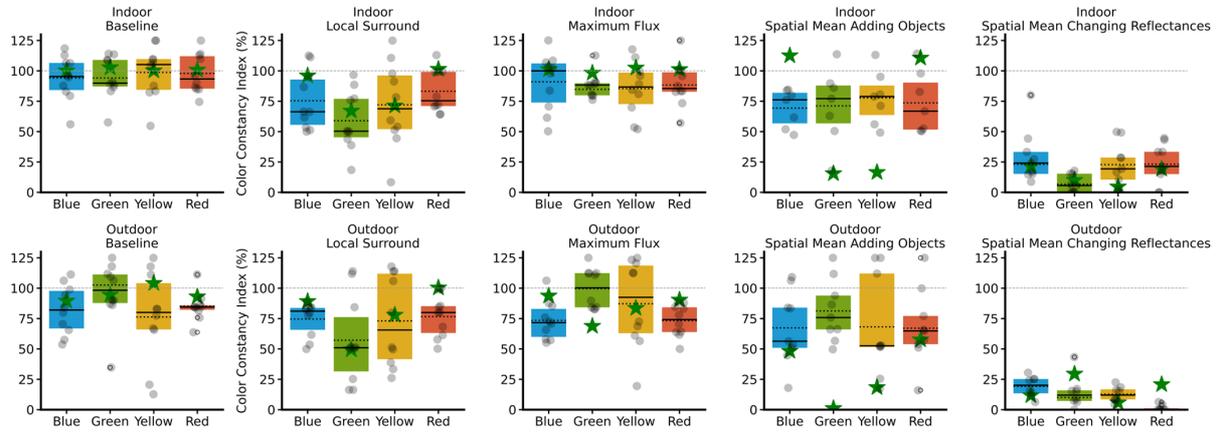

*Figure 8. Color constancy index percentage in all conditions grouped by illuminations. The top row displays data for the indoor environment, and the bottom row for the outdoor environment, with each column representing a different condition. Each gray dot represents a single subject, while the green star shows the model's output. In both panels, the dotted line indicates the mean, and the solid line indicates the median. The horizontal dashed line at 100% represents perfect constancy.*

## 3.2 Experiment 2: Interaction between local surround color and illumination

In the psychophysical experiments, local surround color suppression reduced the model's color constancy performance in both scenes (Table 2). In the indoor scene, baseline CCI values were generally high. In the outdoor scene, baseline performance was lower and more variable. Suppression led to moderate decreases in the indoor scene, while in the outdoor scene, it produced substantially larger reductions, with some conditions dropping close to zero. These results suggest that the local surround plays a more critical role in outdoor scenes, where other available cues may provide less stable information.

*Table 2. Color Constancy Index (CCI) values across illuminant and surface conditions before and after local surround suppression, for indoor and outdoor scenes.*

| Scene | Condition | Surface | Blue | Yellow | Green | Red |
|---|---|---|---|---|---|---|
| **Indoor** | **Baseline** | Khaki | 104.99 | 91.74 | 92.46 | 103.03 |
| | | Rose | 103.92 | 89.78 | 91.29 | 102.69 |
| | | Purple | 112.90 | 80.29 | 82.89 | 103.45 |
| | | Teal | 112.93 | 84.86 | 90.60 | 105.32 |
| | **Suppressed** | Khaki | 101.89 | 87.35 | 101.21 | 111.82 |
| | | Rose | 93.55 | 80.30 | 77.71 | 104.34 |
| | | Purple | 99.55 | 72.67 | 76.45 | 102.53 |
| | | Teal | 105.81 | 77.25 | 90.67 | 113.90 |
| **Outdoor** | **Baseline** | Khaki | 75.55 | 59.24 | 64.29 | 69.04 |

|  |  |  | Rose | 87.07 | 60.11 | 36.00 | 88.32 |
|  |  |  | Purple | 110.36 | 44.44 | 33.29 | 104.49 |
|  |  |  | Teal | 104.41 | 44.55 | 52.24 | 82.60 |
|  | **Suppressed** |  | Khaki | 55.59 | 34.98 | 45.42 | 51.49 |
|  |  |  | Rose | 61.94 | 35.88 | 2.06 | 62.79 |
|  |  |  | Purple | 87.18 | 2.24 | 7.62 | 77.32 |
|  |  |  | Teal | 76.11 | 7.57 | 27.32 | 63.21 |

Figure 11 displays the CCI for indoor and outdoor scenes, highlighting each of the local surround colors, in the x-axis. For each color, the baseline and the local surround suppression mechanism are presented side by side.

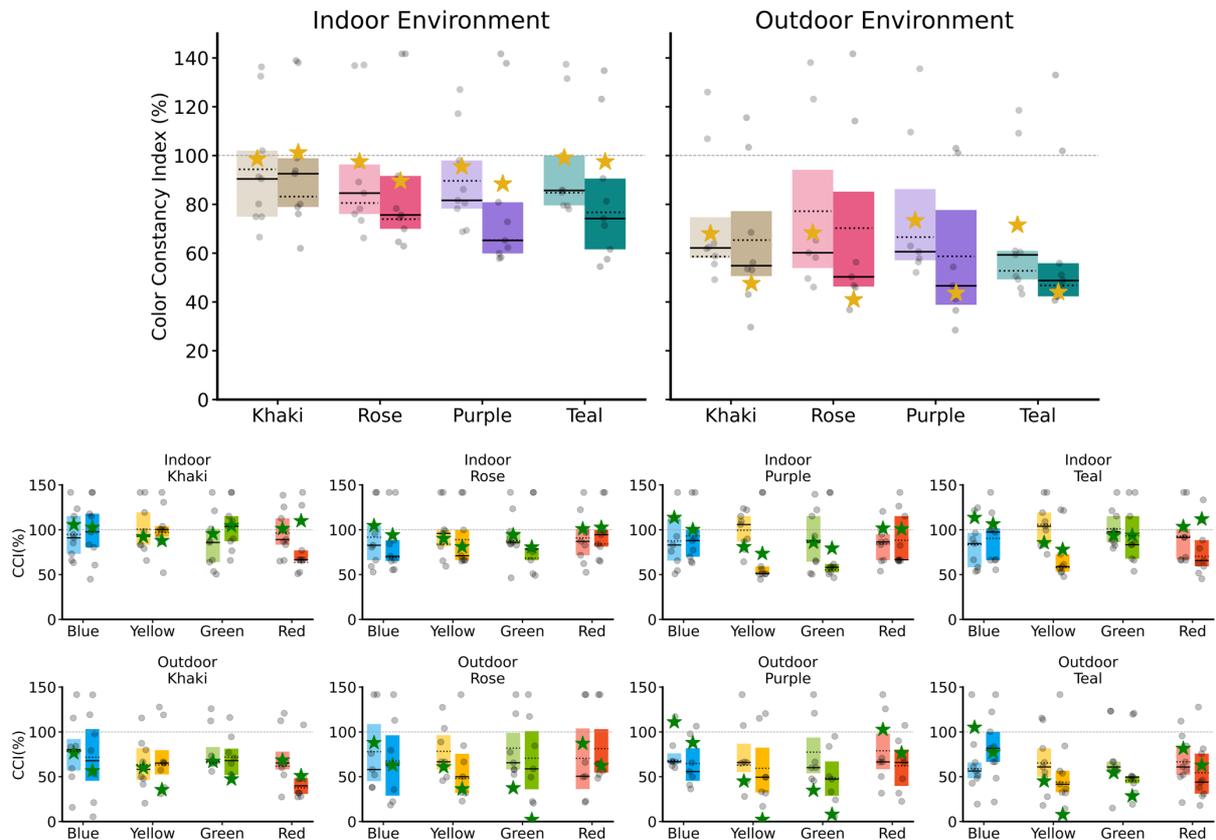

*Figure 11. Color Constancy Index (CCI) results for Experiment 2, comparing human participant data with model predictions. In all plots, the box plots are derived from human data, and each gray dot represents an individual subject. The stars represent the mean CCI of the model for each condition. (Top) Box plots show the overall CCI for indoor (left) and outdoor (right) environments. For each of the four surround colors (Khaki, Rose, Purple, and Teal), the baseline condition (lighter shade) is compared with the local surround suppression condition (darker shade). (Bottom) The CCI is broken down by illuminant color for indoor (upper row) and outdoor (lower row) scenes.*

We aim to investigate how local surround suppression under different color conditions affects performance. To do this, we calculate the difference in CCI, ΔCCI, by subtracting the baseline condition from the experimental condition, as done in the previous experiment. We consider the GLME model of the difference in responses, considering only participants as random effects, with the model considered an additional participant. The average estimate from the model in the indoor scene is -5.638, with a p-value of 0.0135 and a confidence interval of [-10.093, -1.182]. This indicates a significant drop in performance that is statistically significant. In the case of the outdoor scene, we get –10.332 as average, with p-value < 0.0001 and CI of [-15.128, -5.536], also indicating a significant drop in performance.

We can then analyze each color surround separately. In the indoor scene, by applying a GLME model to the ΔCCI values, with local surround color as fixed effects and participants as random effects, we obtain the following results: khaki ($\mu$ = 1.230, p-value=0. 736 and CI [-5.951, 8.410]), purple ($\mu$ = -7.737, p-value=0. 035 and CI [-14.918, -0.557]), rose ($\mu$ = -6.956, p-value=0. 058 and CI [-14.136, 0.225]), and teal ($\mu$ =-9.087, p-value=0. 013 and CI [-16.268, -1.907]). If we do the same for the outdoor scene, khaki ($\mu$ = -8.981, p-value=0. 019 and CI [-16.456,-1.502]), purple ($\mu$ = -18.771, p-value<0.0001 and CI [-26.678,-10.865]), rose ($\mu$ = -9.075, p-value=0. 025 and CI [-16.981,-1.168]), and teal ($\mu$ =-6.233 p-value=0. 085 and CI [-13.336,0.869]).

To study the interaction between the color of the local surround and the color of the illuminant, we consider a GLME model that includes the interaction between both as a fixed effect, with participants as random effects. We then apply ANOVA, resulting in $D(9,144) = 3.905$ and a p-value < 0.001 for the interaction factor. We find that the mean change in ΔCCI for the purple surround color is 3.55 for blue, -12.39 for green, 3.71 for red, and -25.82 for yellow. This indicates that under color illuminants that are closer to the surround color—specifically blue and red—the ΔCCI values are higher compared to those for illuminants that are further away—green and yellow. To explore whether this pattern holds true regardless of the color surround considered, we introduce a new factor called "Direction." In this context, illuminant colors are classified as either Neighboring or Opposing, depending on whether they are close to the local surround color or further away in a contrasting direction. We then fitted a GLME model with "Direction" as a fixed effect and "participants" as random effects. Our analysis revealed that the Neighboring direction was not statistically different from 0, with a p-value of 0.660. In contrast, the Opposing direction was statistically significantly different from 0, showing a p-value of less than 0.0001. Left side of Figure 10 shows ΔCCI that for all participants considering Neighboring and Opposing directions, regardless of the local surround color.

We do the same analysis for the outdoor scene got that for the Neighboring direction ΔCCI are not statistically different from 0, with p-value=0.110, in contrast with Opposing direction were differences in CCI are statistically different from 0 with p-value<0.0001. Right side of Figure 10 shows ΔCCI for the two directions in the outdoor environment.

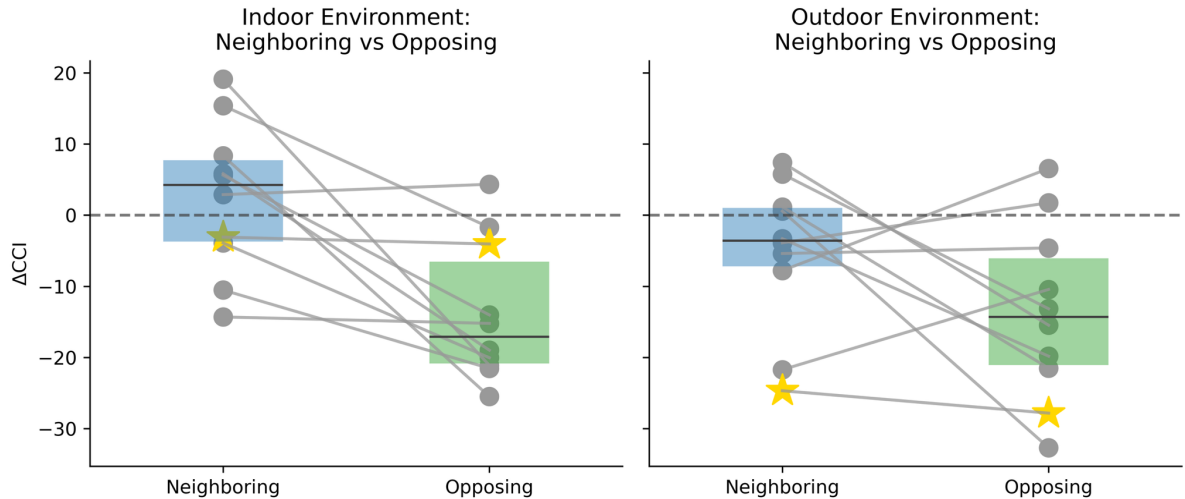

*Figure 10. Change in color constancy performance (ΔCCI) relative to the baseline as a function of chromatic direction. The plots display the difference in CCI for Indoor (left) and Outdoor (right) environments. Conditions are classified as Neighboring (illuminant hue similar to the surround) or Opposing (illuminant hue contrasting with the surround). Gray dots represent individual participants connected by lines to visualize within-subject trends, while the yellow star indicates the model's performance. The boxplots summarize human data: the solid line marks the median and the dotted line indicates the mean.*

## 3.3 Model comparison and validation against human performance

To assess whether the observed effects reflect genuinely human-like cue integration rather than properties of the task or stimulus statistics alone, we compared our model against a range of established color constancy algorithms. These included a simple gray-world baseline that relies on global image statistics, as well as more sophisticated models that incorporate spatial or chromatic normalization. The goal of this comparison was not to identify the model with the highest average CCI, but to evaluate whether different models reproduce the pattern of human performance across experimental manipulations—specifically, the effects of surround silencing, illuminant–surround relationships, and scene context. This comparison therefore provides a critical test of whether the proposed model captures the structure of human color constancy behavior observed in our experiments. Under different illuminants, our model's performance (red-orange bar) consistently tracks the human baseline (green bar), unlike other models that can significantly over- or under-correct relative to human vision (Figure 11A).

This similarity extends to challenging cue-suppression conditions, where our model shares dependencies on visual cues with the human visual system. The fact that when human performance deteriorates, our model also deteriorates indicates an even more mechanistic similarity (Figure 11B). We have quantified this similarity in Figure 11C by showing that the distribution of performance error for our model in relation to human baseline performance (ΔCCI) is tightly centered around zero. The median error is nearly zero, and the variance is substantially smaller than other models, which shows our model provided a more consistent and perceptually accurate description of human color constancy than the other evaluated alternatives.

Finally, to synthesize the overall model-human agreement into a single, comprehensive metric, we calculated the Normalized Lin's Concordance Correlation Coefficient (ncCCC), shown in Figure 11D. This metric quantifies the agreement between each model's predictions and the mean human responses across all conditions. The score is normalized by the human reliability ceiling—a measure of human inter-observer agreement, which we calculated to be 0.931—placing the results on a scale where 1.0 signifies perfect human-level consistency. The results clearly indicate our model's superior performance, achieving an ncCCC score of 0.76. This is substantially higher than the next-best classical algorithms (Shadow of Gray and Weighted Gray Edge, ncCCC ≈ 0.57) and far surpasses traditional methods like White Patch, which showed almost no meaningful agreement with human perception (ncCCC = 0.02).

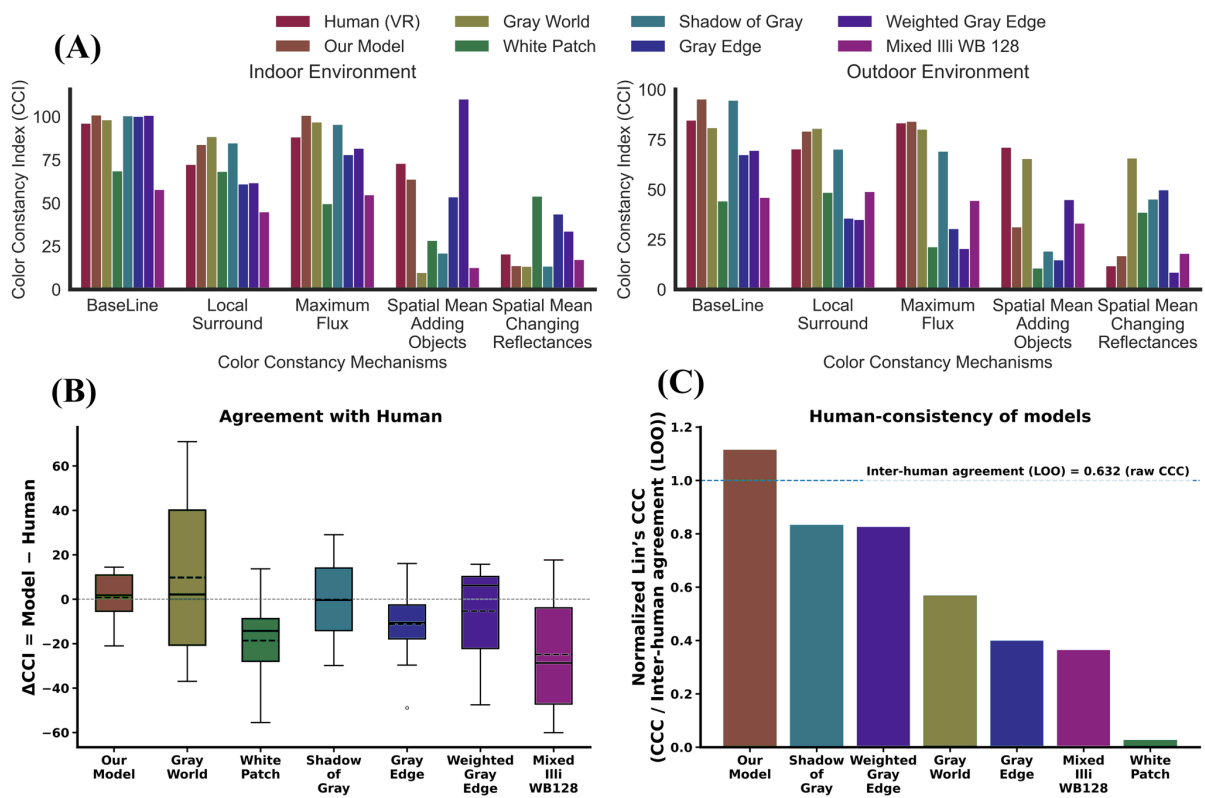

*Figure 11. Comparison of performance metrics for seven color constancy models and a human baseline across various experimental conditions. (A) The Color Constancy Index (CCI) values are presented for each model under a baseline condition and five conditions where specific constancy cues were suppressed: local surround, maximum flux, and spatial mean (disrupted by adding objects or changing reflectances). (B) Box plots illustrate the distribution of the difference in CCI (ΔCCI) between each computational model and the human (VR) data. The central line of each box indicates the median difference, and the box represents the interquartile range. (C) Human consistency of the models, measured as the Normalized Lin's Concordance Correlation Coefficient (ncCCC). This metric is the model-human agreement (CCC) divided by the leave-one-out inter-human agreement (LOO), where the dashed line at 1.0 represents the average level of agreement among humans.*

Table 1 summarizes the signed percentage differences in Color Constancy Index (ΔCCI%) between each experimental condition and its corresponding baseline, for both human data and computational models. Across indoor and outdoor scenes, all mechanisms produced negative ΔCCI% values, indicating a reduction in performance relative to the baseline. The largest drops occurred under the Spatial Mean (changing reflectances) condition, where human observers showed a ΔCCI% of –75.7 % indoors and –72.9 % outdoors, closely followed by our model (–87.1 % and –78.4 %, respectively). This pattern suggests that disrupting global adaptation cues most strongly impairs constancy. The Spatial Mean (adding objects) condition also led to substantial declines for most models, particularly *Weighted Gray Edge* (–79.5 % indoor) and *Shadow of Gray* (–63.9 % outdoor). In contrast, the Local Surround and Maximum Flux mechanisms had milder effects overall, with human ΔCCI% ranging between –8 % and –24 %, and similar trends observed for our model (–0.1 % to –17 %). Among classical algorithms, *Gray World* and *Mixed Illuminant* consistently underperformed across all mechanisms, while *White Patch* exhibited minimal or inconsistent changes, occasionally even positive values in the outdoor scene.

*Table 3. Indoor scene: t*he percentage signed difference between each method's conditions and its own Baseline *(ΔCCI% )*. The CCI of each condition minus the CCI of the Baseline.

|  |  | Human data | Our model | Shadow of gray | Weighted gray edge | Gray world | Gray edge | Mixed illum | White Patch |
|---|---|---|---|---|---|---|---|---|---|
| IN | Local Surround | -23.801 | -17.040 | -15.704 | -38.960 | -9.768 | -39.111 | -12.870 | -0.274 |
|  | Maximum Flux | -8.009 | -0.141 | -4.973 | -18.992 | -1.288 | -22.129 | -3.108 | -18.908 |
|  | SM adding | -23.260 | -37.141 | -79.495 | 9.470 | -88.304 | -46.520 | -45.218 | -40.163 |
|  | SM changing | -75.703 | -87.137 | -87.009 | -66.988 | -84.755 | -56.484 | -40.518 | -14.539 |
| Out | Local Surround | -14.42 | -16.06 | -24.45 | -34.63 | -0.33 | -31.83 | 2.95 | 4.34 |
|  | Maximum Flux | -1.39 | -11.20 | -25.49 | -49.12 | -0.69 | -37.03 | -1.50 | -22.84 |
|  | SM adding | -13.61 | -63.99 | -75.47 | -24.66 | -15.38 | -52.66 | -12.86 | -33.49 |
|  | SM changing | -72.93 | -78.39 | -49.44 | -60.97 | -15.17 | -17.62 | -27.93 | -5.63 |
| All | Local Surround | -19.11 | -16.55 | -20.08 | -36.80 | -5.05 | -35.47 | -4.96 | 2.03 |
|  | Maximum Flux | -4.70 | -5.67 | -15.23 | -34.06 | -0.99 | -29.58 | -2.30 | -20.88 |
|  | SM adding | -18.43 | -50.57 | -77.48 | -7.60 | -51.84 | -49.59 | -29.04 | -36.83 |
|  | SM changing | -74.32 | -82.76 | -68.22 | -63.98 | -49.96 | -37.05 | -34.23 | -10.08 |

Tables 4 reports the accuracy, normalized error, and bias of the models CCI relative to the human data. Accuracy is defined as the Pearson correlation between model performance and the mean human performance across conditions. Bias is defined as the mean of the difference between the model prediction and the mean human value, while the normalized error is computed as the RMSE normalized by the standard deviation of the human data. Metrics are reported separately for indoor scenes, outdoor scenes, and for all conditions combined.

In Table 5, the same analysis is performed on CCI values after baseline correction, obtained by subtracting the Baseline CCI from each condition for both the models and the human data.

*Table 4. Accuracy, normalized error and bias of each of the models with respect to human data CCI.*

|  |  | Our model | Shadow of gray | Weighted gray edge | Gray world | Gray edge | Mixed illum | White Patch |
|---|---|---|---|---|---|---|---|---|
| All | Accuracy | 0.552 | 0.410 | 0.458 | 0.071 | .347 | 0.206 | 0.382 |
|  | N. Error | 1.144 | 1.308 | 1.247 | 1.795 | 1.292 | 1.759 | 1.356 |
|  | Bias | -0.170 | -5.802 | -10.476 | 0.794 | -13.685 | -29.304 | -23.924 |
| In | Accuracy | 0.793 | 0.761 | 0.770 | 0.667 | 0.589 | 0.471 | 0.077 |
|  | N. Error | 0.888 | 0.948 | 0.750 | 1.170 | 0.928 | 1.616 | 1.308 |
|  | Bias | 2.597 | -6.990 | 7.569 | -8.710 | -2.746 | -32.516 | -16.278 |
| Out | Accuracy | 0.676 | 0.348 | 0.485 | 0.188 | -0.089 | 0.231 | -0.095 |
|  | N. Error | 0.905 | 1.294 | 1.386 | 1.708 | 1.627 | 1.687 | 1.766 |
|  | Bias | -2.938 | -4.613 | -28.521 | 10.298 | -24.624 | -26.092 | -31.570 |

*Table 5. Accuracy, normalized error and bias of each of the models with respect to human data based on the ΔCCI.*

|  |  | Our model | Shadow of gray | Weighted gray edge | Gray world | Gray edge | Mixed illum | White Patch |
|---|---|---|---|---|---|---|---|---|
| All | Accuracy | 0.521 | 0.387 | 0.263 | -0.075 | 0.110 | -0.009 | 0.202 |
|  | N. Error | 1.214 | 1.320 | 1.275 | 1.646 | 1.323 | 1.399 | 1.158 |
|  | Bias | -7.797 | -12.891 | -5.173 | 1.744 | -7.025 | 9.207 | 10.162 |
| In | Accuracy | 0.797 | 0.777 | 0.755 | 0.712 | 0.621 | 0.443 | 0.126 |
|  | N. Error | 0.890 | 0.974 | 0.765 | 1.132 | 0.924 | 1.035 | 1.198 |
|  | Bias | -2.137 | -11.282 | 3.061 | -10.668 | -6.694 | 5.812 | 11.378 |
| Out | Accuracy | 0.664 | 0.371 | 0.508 | 0.358 | 0.033 | 0.309 | -0.109 |
|  | N. Error | 1.036 | 1.224 | 1.076 | 1.050 | 1.284 | 1.158 | 1.267 |
|  | Bias | -13.457 | -14.500 | -13.407 | 14.156 | -7.356 | 12.602 | 8.947 |

Across all conditions, our model showed the highest accuracy and smallest bias relative to human data (see Tables 2, 3, and 4). Indoor conditions yielded higher accuracy and smaller normalized error across all models compared to outdoor conditions. Classical algorithms such as Gray World, Gray Edge, and White Patch exhibited lower accuracy and pronounced biases, especially under outdoor illumination. Our model, along with Shadow of Gray and Weighted Gray Edge, maintained moderate-to-high accuracy indoors (r ≈ 0.75–0.80), while traditional single-illuminant models exhibited strong negative biases outdoors. Gray Edge and Mixed Illumination models occasionally produced near-zero or even positive biases, but these appeared unstable and scene-dependent. Across models, substantial differences emerged in how color constancy varied across experimental conditions. While some models achieved moderate to high average CCI values, most failed to reproduce the selective degradation observed in human observers when local surround information was removed or when illuminants were chromatically opposing the surround. In particular, the gray-world baseline showed little sensitivity to surround manipulation, indicating that global image statistics alone are insufficient to account for the human data. By contrast, our model most consistently captured both the robustness of constancy under baseline conditions and the structured, direction-dependent decline under surround silencing, closely paralleling the human results. These findings demonstrate that matching overall constancy performance is not sufficient; reproducing the relative effects of cue availability and chromatic context is critical for explaining human color constancy.

# 4. Discussion

Using human performance as a behavioral reference, the present work evaluates how different computational models of color constancy respond to systematic manipulations of color constancy mechanisms, including local surround, spatial mean, and maximum flux, under different illuminants. The results show that models differ markedly in their ability to reproduce the structure of human responses across conditions, even when they achieve comparable levels of overall constancy. Models based primarily on global image statistics exhibit limited sensitivity to surround manipulations and fail to capture the selective performance changes observed under surround silencing. By contrast, the proposed model reproduces both the robustness of human performance when multiple cues are available and the structured degradation that emerges when local surround information is removed. These findings indicate that capturing human-like color constancy requires more than estimating an average illuminant; it requires mechanisms that integrate local chromatic context with global scene information in a condition-dependent manner.

## 4.1 Model comparison against human performance

To understand why different models succeed or fail under these conditions, we interpret their behavior in terms of the specific mechanisms they implement. This approach shifts the emphasis from ranking models by aggregate performance to identifying which classes of computational strategies are necessary to reproduce the structured, condition-dependent patterns observed in human color constancy.

### 4.1.1 Spatial Mean Mechanism and the Failure of Global Averaging

The "Gray World" hypothesis assumes that the average color of a scene is achromatic (Buchsbaum, 1980). Our results demonstrate that models relying strictly on this assumption, specifically the Gray

World and Shadow of Gray algorithms, fail to replicate human performance when the spatial mean is explicitly manipulated.

In the indoor scene, both algorithms suffered significant performance drops when the scene mean was biased (via added objects or changed reflectances), mirroring the expected violation of their core assumption. However, in the outdoor scene, the Gray World algorithm maintained unexpectedly high CCI values under specific illuminants. Analysis of the input images reveals that this robustness is an artifact of the scene content rather than algorithmic stability. As confirmed by colorimetric measurements of the outdoor scene (Figure 12), the presence of grass and trees introduces a strong green shift in the global mean (Rodríguez et al., 2024). This shift coincidentally favors the Gray World mechanism under green and yellow illuminants, resulting in high CCI values (approx. 117–123%), while performance under blue and red illuminants remains poor (< 15%). Consequently, while the average performance appears high, the algorithm fails to generalize across chromatic conditions, diverging from the more stable human pattern.

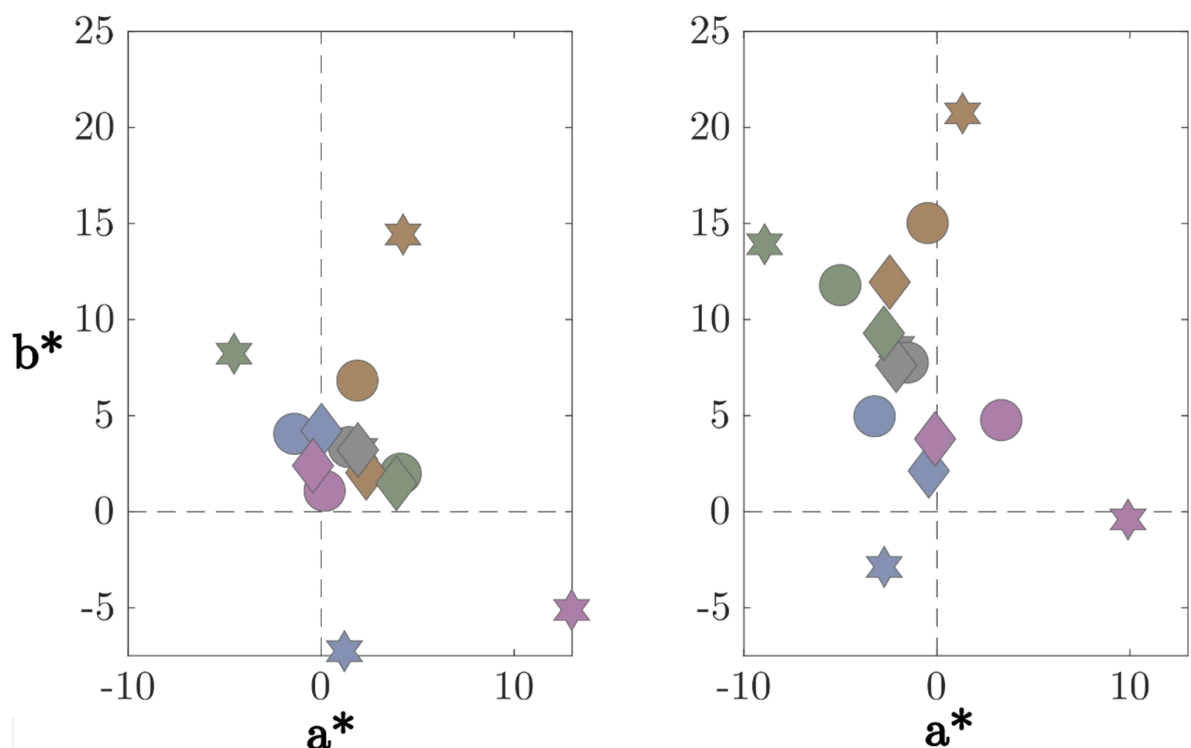

*Figure 12. Mean color of the scenes plotted on the CIELAB a\*–b\* plane. The symbols indicate the mechanism: Baseline (hexagon), Spatial Mean with added objects (circle), and Spatial Mean with varying reflectances (diamond). The marker colors correspond to the five illumination conditions. Results are shown for Indoor (left) and Outdoor (right) environments.*

### 4.1.2 Maximum Flux Mechanism and its Redundancy via Alternative Cues

The White Patch algorithm (Land, 1977), which estimates the illuminant from the brightest region of an image, showed surprising robustness in the "Maximum Flux" suppression condition. In principle, suppressing the maximum flux cue should impair this algorithm. However, its performance did not drop significantly in either Indoor or Outdoor scenes. This can likely be attributed to the sampling procedure. Model evaluation was based on multiple screenshots covering different regions of the scene. Because the manipulated white region (the shaded lamp in the Indoor scene and the paper or clothing on the cliff in the Outdoor scene) occupied only a small portion of the visual field, it was absent from many samples. As a result, the White Patch algorithm retained a bright pixel in most inputs, maintaining good overall performance despite the local manipulation. Importantly, this sampling procedure may approximate human viewing behavior, where observers also sample scenes sequentially rather than relying on a single global snapshot, which could explain why the manipulation had limited behavioral impact.

This finding highlights a critical distinction between heuristic and human-like cue use. Human observers appear to integrate maximum flux information as one of several weighted cues (Foster, 2011), whereas the White Patch algorithm remains robust as long as any sufficiently bright pixel is present, without flexible cue integration. By contrast, the CNN-based Mixed Illuminant model showed limited sensitivity to these manipulations, suggesting that high-intensity regions are not treated as uniquely diagnostic features in its learned representation. Similarly, algorithms relying primarily on global image statistics exhibited only moderate performance changes under this condition, reflecting their relative insensitivity to localized high-luminance regions.

### 4.1.3 Local Surround Mechanism, Edge Dependency and Contextual Integration

The Gray Edge and Weighted Gray Edge algorithms (Van De Weijer et al., 2007) derive illuminant estimates from the average color difference at edges. These methods exhibited a significant performance drop when the local surround was manipulated. Since these algorithms rely heavily on local edges (e.g., the boundaries of leaves surrounding the target), altering the surround color directly corrupted their primary input data. This error was consistent across most illuminants, with the exception of the red illuminant. In the red condition, the surround color was chromatically close to the global scene illumination, allowing the edge-based estimate to remain accurate by chance. In contrast, our proposed model demonstrated a pattern of degradation that closely mirrored human observers. Unlike the edge-based algorithms, which failed whenever local edges were misleading, our model's performance depended on the chromatic relationship between the illuminant and the surround. This suggests that the model does not merely aggregate edge statistics but learns a context-dependent representation of the surround. This behavior aligns with recent computational evidence suggesting that accurate color constancy models should mimic human-like susceptibility to contextual illusions, such as color assimilation, rather than treating local context merely as noise to be filtered (Ulucan et al., 2026). Other global statistics-based algorithms, such as Gray World and White Patch, showed relatively stable performance under this condition, as they do not rely on local edge information and were therefore unaffected by the surround manipulation.

### 4.1.4 Interpretation of Model Differences

The quantitative comparison highlights that models incorporating spatial weighting or local context better align with human color constancy than those relying solely on global image statistics. Our model's ability to produce lower bias and lower normalized error aligns with research suggesting that computational models remain some distance from human-like constancy under naturalistic conditions (Foster, 2011). The indoor-outdoor performance gap further underscores the importance of contextual information and spatial structure in predicting human-like constancy. These findings extend the benchmark results of Gijsenij et al. (2011), where Gray World and White Patch exhibited limited robustness to complex illumination changes. The consistency of our model's performance across both scene types supports its robustness and indicates that it captures key perceptual mechanisms underlying human adaptation to complex illumination.

## 4.2 Interaction between Local Surround Color and Illumination

Analysis of the local surround mechanism reveals a clear distinction between indoor and outdoor scenes. In the indoor environment, the model consistently exhibits high CCI values under the red illuminant, regardless of the local surround color (khaki, rose, purple, teal). This suggests that it can recover the target reflectance effectively even when local contrast cues are altered. In contrast, in the outdoor scene, CCI under the red illuminant is more variable and generally lower, particularly for surrounds that differ strongly from red. This difference likely arises from the greater spatial and chromatic complexity of the outdoor environment, including shadows, indirect lighting, and interactions between multiple surfaces, which challenge the model's estimation. The simpler and more uniform indoor scene allows the model to rely on global or central color cues, while the outdoor scene requires handling more diverse chromatic information. In addition, the model's architecture (U-Net with PBCLoss) and training data may favor indoor-type illumination conditions, and the red illuminant may correspond to hues the model has higher confidence in, stabilizing indoor predictions but producing less robust results outdoors.

Color-specific analyses indicate that purple and teal surrounds significantly reduce performance indoors, while purple, khaki, and rose are significant outdoors, highlighting the influence of surround hue on target color perception. Interaction analyses show that ΔCCI declines most when the illuminant is chromatically opposite to the surround ("Opposing"), whereas "Neighboring" illuminants produce no significant effect. This pattern was consistent across both scenes, though the magnitude of the effect was notably larger in the outdoor environment (see Figure 10).

Comparing the two experiments further clarifies the effect of baseline definition. In Experiment 1, the Local Surround mechanism showed a significant performance drop for rose-colored leaves in both indoor and outdoor scenarios, with a more pronounced decrease indoors. In Experiment 2, the drop for rose leaves was greater outdoors, a pattern generally observed across nearly all surround colors. The main difference between experiments was that the Baseline in Experiment 2 included a leaf beneath the targets, introducing additional local contrast that interacted with illuminant color and altered the perceived target color. Overall, these findings indicate that while the model replicates human-like local surround effects, its performance depends on scene complexity, illuminant-surround interactions, and baseline definitions, highlighting both its strengths and the contexts in which perceptual limitations emerge.

## 4.3  Why the Proposed Model Aligns with Human Performance

A fundamental paradox in color constancy research is that human observers exhibit substantial constancy for surface colors while simultaneously showing poor accuracy in estimating illuminant properties (Smithson, 2005; Granzier et al., 2009; Radonjić et al., 2015). When asked to judge the color of ambient illumination directly, observers produce highly variable and often inaccurate estimates (Rutherford and Brainard, 2002; Reeves et al., 2008). Yet these same observers reliably identify object colors across illumination changes. This dissociation suggests that human color constancy may not require, and perhaps does not involve, explicit illuminant recovery (Foster, 2011; Witzel and Gegenfurtner, 2018).

This finding poses a challenge for the dominant computational paradigm. The majority of algorithms, from classical Gray World and White Patch to modern deep learning approaches (Bianco et al., 2017; Hu et al., 2017; Yu et al., 2020; Afifi et al., 2021; Lo et al., 2021; Domislović et al., 2022; Buzzelli et al., 2023), formulate color constancy as a two-stage process that first estimates the scene illuminant and then applies a chromatic adaptation transform (Land, 1977; Buchsbaum, 1980; Gijsenij et al., 2011; Hu et al., 2017). While these illumination-centric models achieve reasonable benchmark performance, their underlying assumption appears to conflict with psychophysical evidence. Granzier and Gegenfurtner (Granzier and Gegenfurtner, 2012) demonstrated that observers maintained color constancy for familiar objects even when they could not accurately report illuminant chromaticity. Pearce et al. (Pearce et al., 2014) showed that observers' illuminant estimates were biased and variable, yet their color constancy remained robust.

If not through illuminant estimation, how does the human visual system achieve constancy? Substantial evidence points to relational processing, defined as the encoding of chromatic relationships between surfaces rather than absolute color values ((Land and McCann, 1971; Gilchrist et al., 1999). Land's Retinex theory (Land, 1983) proposed that lightness is computed from ratios of intensities across edges rather than from absolute values. Foster and Nascimento (Foster and Nascimento, 1994; Nascimento and Foster, 2000) extended this to chromatic processing, demonstrating that cone-excitation ratios across surface boundaries remain approximately invariant under illumination changes. Importantly, relational processing may not require explicit knowledge of the illuminant. By encoding how surfaces relate to one another chromatically, the visual system could maintain stable representations without solving the inverse problem of illuminant estimation.

Our model takes a different approach from illumination-estimation methods. Rather than estimating scene illumination and applying chromatic correction, the model directly predicts surface reflectance values at each pixel (Heidari-Gorji and Gegenfurtner, 2023). This formulation aligns with how color constancy is defined in human vision research, which emphasizes the stability of perceived surface color, not the accuracy of illuminant recovery. The U-Net architecture (Ronneberger et al., 2015) enables this direct prediction by combining information at multiple spatial scales through skip connections, allowing local chromatic relationships to inform the output alongside global scene context. This multi-scale integration mirrors theoretical accounts emphasizing the combination of local relational cues with spatial mean adaptation (Kraft and Brainard, 1999; Hansen et al., 2007; Ulucan et al., 2026).

During training on rendered scenes with ground-truth reflectances, the network learns which image features predict stable surface colors across illumination changes. The features it discovers likely include relational information such as edge contrasts, chromatic ratios, and spatial context, because these cues remain informative when illumination varies. This emergent relational processing may parallel the strategy employed by human vision. The Perceptual Balanced Color Loss (PBCLoss) further aligns optimization with human perception by combining CIEDE2000 (Sharma et al., 2005), which provides perceptually uniform error measurement, with chroma-weighted terms that prevent the model from learning trivial desaturated solutions.

This theoretical framework may help explain why our model reproduces human performance patterns while illumination-estimation models do not. When scene reflectances are uniformly altered to oppose the illuminant (spatial mean manipulation), both humans and our model show severe constancy failures. An illumination-estimation model might maintain performance if it can still extract illuminant information from unchanged regions. But if our model, possibly like human vision, relies on chromatic relationships between surfaces, altering those relationships would disrupt the cues on which constancy depends. Similarly, local surround manipulation selectively impairs both humans and our model when illuminant and surround are chromatically opposing, because relational processing depends on edge chromatic structure. The Mixed Illuminant CNN, trained for illuminant estimation, showed limited sensitivity to these manipulations since it extracts illuminant-diagnostic features that generalize poorly when some cues are preserved while others are disrupted. Notably, recent computational work has also moved beyond pure illuminant estimation, exploring unified frameworks that address both color constancy and color illusion phenomena (Ulucan et al., 2024), and investigating the emergence of human-like color invariance in networks trained for image segmentation (Hernández-Cámara et al., 2024). Our findings suggest that the illumination-estimation paradigm may not fully capture human perception. Models achieving constancy through illuminant recovery may solve a different computational problem than the human visual system. The success of direct reflectance prediction supports an alternative framework that views color constancy as relational surface color encoding, consistent with decades of psychophysical research (Land and McCann, 1971; Foster and Nascimento, 1994; Gilchrist et al., 1999; Nascimento and Foster, 2000) and recent efforts to unify color constancy and visual illusions within a single computational framework (Ulucan et al., 2026). By training networks to predict surface colors directly rather than through illuminant intermediaries, we obtain models that achieve constancy in a human-like manner.

## Conclusion

In this study, we investigated how a DNN model trained to estimate pixel-wise reflectance of an input image replicates human performance in a psychophysical color constancy experiment using a selection task paradigm. The original experiment by Gil Rodríguez et al. (2024) examined well-known color constancy mechanisms—local surround, maximum flux, and spatial mean—comparing performance against a baseline condition to quantify the influence of each mechanism. The experiment was conducted in VR across two distinct scenes.

To adapt the model to these new data, we applied transfer learning, training only the decoder using the baseline condition for both indoor and outdoor scenes. Subsequently, screenshots representing each mechanism with various target colors were used to assign a CCI to the model under each condition and illuminant. The results show that the model performs nearly perfectly under the baseline condition and closely mirrors human performance across most mechanisms. The notable exception is the Spatial Mean mechanism with added objects, where the model underperforms relative to humans.

We also compared the model's performance to human data for the local surround mechanism with different surround colors(Rodriguez et al., 2024). In the indoor scene, human performance is influenced by both the local surround color and the chromatic direction of the illuminant, whereas the model's performance remains largely unaffected. In contrast, in the outdoor scene, both humans and the model exhibit a clear effect of chromatic direction on performance, demonstrating a closer alignment in more complex environments.

Finally, we benchmarked our model against classical and state-of-the-art color constancy algorithms, including Gray World, White Patch, Shadow of Gray, Gray Edge, Weighted Gray Edge, and a mixed-illuminant CNN model. Our model most closely replicates human data, outperforming these alternative approaches. Understanding the underlying reasons for the model's behavior may provide further insights into human color perception.

# Acknowledgments

This work was supported by European Research Council ERC AdG Color 3.0 (884116) and by DFG Sonderforschungsbereich SFB TRR 135 project C2 (222641018).

# References

Afifi, M., Barron, J. T., LeGendre, C., Tsai, Y.-T., and Bleibel, F. (2021). Cross-Camera Convolutional Color Constancy., in *2021 IEEE/CVF International Conference on Computer Vision (ICCV)*, 1961–1970. doi: 10.1109/ICCV48922.2021.00199

Afifi, M., Brubaker, M. A., and Brown, M. S. (2022). Auto White-Balance Correction for Mixed-Illuminant Scenes., 1210–1219.

Aston, S., Radonjić, A., Brainard, D. H., and Hurlbert, A. C. (2019). Illumination discrimination for chromatically biased illuminations: Implications for color constancy. *Journal of Vision* 19, 15. doi: 10.1167/19.3.15

Bianco, S., Cusano, C., and Schettini, R. (2015). Color Constancy Using CNNs. doi: 10.48550/arXiv.1504.04548

Bianco, S., Cusano, C., and Schettini, R. (2017). Single and Multiple Illuminant Estimation Using Convolutional Neural Networks. *IEEE Transactions on Image Processing* 26, 4347–4362. doi: 10.1109/TIP.2017.2713044

Brainard, D. H., and Maloney, L. T. (2011). Surface color perception and equivalent illumination models. *Journal of vision* 11, 1–1.

Brainard, D. H., and Wandell, B. A. (1986). Analysis of the retinex theory of color vision. *Journal of the Optical Society of America A* 3, 1651–1661.

Buchsbaum, G. (1980). A spatial processor model for object colour perception. *Journal of the Franklin Institute* 310, 1–26. doi: 10.1016/0016-0032(80)90058-7

Buzzelli, M., Schettini, R., and Bianco, S. (2023). Learning color constancy: 30 years later., in *Color and Imaging Conference*, (Society for Imaging Science and Technology), 91–95. Available at: https://library.imaging.org/cic/articles/31/1/17 (Accessed February 9, 2026).

Cataliotti, J., and Gilchrist, A. (1995). Local and global processes in surface lightness perception. *Perception & Psychophysics* 57, 125–135. doi: 10.3758/BF03206499

Domislović, I., Vršnak, D., Subašić, M., and Lončarić, S. (2022). One-net: Convolutional color constancy simplified. *Pattern Recognition Letters* 159, 31–37. doi: 10.1016/j.patrec.2022.04.035

Ebner, M. (2007). *Color constancy*. John Wiley & Sons.

Entok, U. C., Laakom, F., Pakdaman, F., and Gabbouj, M. (2024). Pixel-Wise Color Constancy Via Smoothness Techniques In Multi-Illuminant Scenes., in *2024 IEEE International Conference on Image Processing (ICIP)*, 2737–2743. doi: 10.1109/ICIP51287.2024.10647547

Erba, I., Buzzelli, M., Thomas, J.-B., Hardeberg, J. Y., and Schettini, R. (2024). Improving RGB illuminant estimation exploiting spectral average radiance. *J. Opt. Soc. Am. A, JOSAA* 41, 516–526. doi: 10.1364/JOSAA.510159

Finlayson, G. D., and Trezzi, E. (2004). Shades of Gray and Colour Constancy. *Color and Imaging Conference* 12, 37–41. doi: 10.2352/CIC.2004.12.1.art00008

Flachot, A., Akbarinia, A., Schütt, H. H., Fleming, R. W., Wichmann, F. A., and Gegenfurtner, K. R. (2022). Deep neural models for color classification and color constancy. *Journal of Vision* 22, 17. doi: 10.1167/jov.22.4.17

Foster, D. H. (2011). Color constancy. *Vision Research* 51, 674–700. doi: 10.1016/j.visres.2010.09.006

Foster, D. H., and Nascimento, S. M. (1994). Relational colour constancy from invariant cone-excitation ratios. *Proc Biol Sci* 257, 115–121. doi: 10.1098/rspb.1994.0103

Gegenfurtner, K. R., and Kiper, D. C. (2003). Color Vision. *Annual Review of Neuroscience* 26, 181–206. doi: 10.1146/annurev.neuro.26.041002.131116

Gijsenij, A., and Gevers, T. (2007). Color Constancy using Natural Image Statistics., in *2007 IEEE Conference on Computer Vision and Pattern Recognition*, 1–8. doi: 10.1109/CVPR.2007.383206

Gijsenij, A., Gevers, T., and Van De Weijer, J. (2011). Computational color constancy: Survey and experiments. *IEEE transactions on image processing* 20, 2475–2489.

Gil Rodríguez, R., Bayer, F., Toscani, M., Guarnera, D., Guarnera, G. C., and Gegenfurtner, K. R. (2022). Colour Calibration of a Head Mounted Display for Colour Vision Research Using Virtual Reality. *SN COMPUT. SCI.* 3, 22. doi: 10.1007/s42979-021-00855-7

Gilchrist, A., Kossyfidis, C., Bonato, F., Agostini, T., Cataliotti, J., Li, X., et al. (1999). An anchoring theory of lightness perception. *Psychol Rev* 106, 795–834. doi: 10.1037/0033-295x.106.4.795

Gilchrist, A. L., and Radonjić, A. (2010). Functional frameworks of illumination revealed by probe disk technique. *Journal of Vision* 10, 6–6.

Granzier, J. J., Brenner, E., and Smeets, J. B. (2009). Can illumination estimates provide the basis for color constancy? *Journal of Vision* 9, 18–18.

Granzier, J. J. M., and Gegenfurtner, K. R. (2012). Effects of Memory Colour on Colour Constancy for Unknown Coloured Objects. *i-Perception* 3, 190–215. doi: 10.1068/i0461

Hansen, T., Walter, S., and Gegenfurtner, K. R. (2007). Effects of spatial and temporal context on color categories and color constancy. *Journal of Vision* 7, 2. doi: 10.1167/7.4.2

Heidari-Gorji, H., and Gegenfurtner, K. R. (2023). Object-based color constancy in a deep neural network. *J Opt Soc Am A Opt Image Sci Vis* 40, A48–A56. doi: 10.1364/JOSAA.479451

Hernández-Cámara, P., Daudén-Oliver, P., Laparra, V., and Malo, J. (2024). Alignment of color discrimination in humans and image segmentation networks. *Front. Psychol.* 15. doi: 10.3389/fpsyg.2024.1415958

Hu, Y., Wang, B., and Lin, S. (2017). FC^4: Fully Convolutional Color Constancy with Confidence-Weighted Pooling., in *2017 IEEE Conference on Computer Vision and Pattern Recognition (CVPR)*, 330–339. doi: 10.1109/CVPR.2017.43

Hurlbert, A. (2019). Challenges to color constancy in a contemporary light. *Current Opinion in Behavioral Sciences* 30, 186–193. doi: 10.1016/j.cobeha.2019.10.004

Kim, D., Kim, J., Yu, J., and Kim, S. J. (2024). Attentive Illumination Decomposition Model for Multi-Illuminant White Balancing., in *2024 IEEE/CVF Conference on Computer Vision and Pattern Recognition (CVPR)*, 25512–25521. doi: 10.1109/CVPR52733.2024.02410

Kraft, J. M., and Brainard, D. H. (1999). Mechanisms of color constancy under nearly natural viewing. *Proceedings of the National Academy of Sciences* 96, 307–312. doi: 10.1073/pnas.96.1.307

Land, E. H. (1977). The retinex theory of color vision. *Scientific american* 237, 108–129.

Land, E. H. (1983). Recent advances in retinex theory and some implications for cortical computations: color vision and the natural image. *Proc Natl Acad Sci U S A* 80, 5163–5169. doi: 10.1073/pnas.80.16.5163

Land, E. H., and McCann, J. J. (1971). Lightness and Retinex Theory. *J. Opt. Soc. Am., JOSA* 61, 1–11. doi: 10.1364/JOSA.61.000001

Lo, Y.-C., Chang, C.-C., Chiu, H.-C., Huang, Y.-H., Chen, C.-P., Chang, Y.-L., et al. (2021). Clcc: Contrastive learning for color constancy., in *Proceedings of the IEEE/CVF conference on computer vision and pattern recognition*, 8053–8063. Available at: http://openaccess.thecvf.com/content/CVPR2021/html/Lo_CLCC_Contrastive_Learning_for_Color_Constancy_CVPR_2021_paper.html (Accessed February 9, 2026).

Lou, Z., Gevers, T., Hu, N., and Lucassen, M. P. (2015). Color Constancy by Deep Learning., in *BMVC*, 76–1.

Maloney, L. T., and Knoblauch, K. (2020). Measuring and Modeling Visual Appearance. *Annu Rev Vis Sci* 6, 519–537. doi: 10.1146/annurev-vision-030320-041152

Milidonis, X., Artusi, A., and Banterle, F. (2025). Deep Chroma Compression of Tone-Mapped Images. *ACM Trans. Multimedia Comput. Commun. Appl.*, 3744925. doi: 10.1145/3744925

Nascimento, S. M. C., and Foster, D. H. (2000). Relational color constancy in achromatic and isoluminant images. *J. Opt. Soc. Am. A, JOSAA* 17, 225–231. doi: 10.1364/JOSAA.17.000225

Pearce, B., Crichton, S., Mackiewicz, M., Finlayson, G. D., and Hurlbert, A. (2014). Chromatic illumination discrimination ability reveals that human colour constancy is optimised for blue daylight illuminations. *PLoS One* 9, e87989. doi: 10.1371/journal.pone.0087989

Radonjić, A., and Brainard, D. H. (2016). The nature of instructional effects in color constancy. *Journal of Experimental Psychology: Human Perception and Performance* 42, 847.

Radonjić, A., Cottaris, N. P., and Brainard, D. H. (2015). Color constancy in a naturalistic, goal-directed task. *Journal of Vision* 15, 3–3.

Reeves, A. J., Amano, K., and Foster, D. H. (2008). Color constancy: Phenomenal or projective? *Perception & Psychophysics* 70, 219–228. doi: 10.3758/PP.70.2.219

Rodríguez, R. G., Hedjar, L., Toscani, M., Guarnera, D., Guarnera, G. C., and Gegenfurtner, K. R. (2024). Color constancy mechanisms in virtual reality environments. *Journal of Vision* 24, 6–6.

Rodriguez, R. G., Hedjar, L., Toscani, M., Guarnera, D., Guarnera, G. C., and Gegenfurtner, K. R. (2024). Surround effects on color constancy in virtual reality. *Journal of Vision* 24, 517–517.

Ronneberger, O., Fischer, P., and Brox, T. (2015). U-net: Convolutional networks for biomedical image segmentation., in *International Conference on Medical image computing and computer-assisted intervention*, (Springer), 234–241.

Rutherford, M. D., and Brainard, D. H. (2002). Lightness constancy: a direct test of the illumination-estimation hypothesis. *Psychol Sci* 13, 142–149. doi: 10.1111/1467-9280.00426

Sharma, G., Wu, W., and Dalal, E. N. (2005). The CIEDE2000 color-difference formula: Implementation notes, supplementary test data, and mathematical observations. *Color Res. Appl.* 30, 21–30. doi: 10.1002/col.20070

Smithson, H. E. (2005). Sensory, computational and cognitive components of human colour constancy. *Philosophical Transactions of the Royal Society B: Biological Sciences* 360, 1329–1346. doi: 10.1098/rstb.2005.1633

Ulucan, O., Ulucan, D., and Ebner, M. (2022). BIO-CC: Biologically inspired color constancy., in *BMVC*, 960. Available at: https://stubber.math-inf.uni-greifswald.de/~ebner/resources/uniG/BMVC-2022-BioCC.pdf (Accessed February 5, 2026).

Ulucan, O., Ulucan, D., and Ebner, M. (2024). A computational model for color assimilation illusions and color constancy., in *Proceedings of the Asian Conference on Computer Vision*, 630–647. Available at: https://openaccess.thecvf.com/content/ACCV2024/html/Ulucan_A_computational_model_for_color_assimilation_illusions_and_color_constancy_ACCV_2024_paper.html (Accessed February 9, 2026).

Ulucan, O., Ulucan, D., and Ebner, M. (2026). A Traditional Approach for Color Constancy and Color Assimilation Illusions with Its Applications to Low-Light Image Enhancement. *Int J Comput Vis* 134, 33. doi: 10.1007/s11263-025-02595-0

Valberg, A., and Lange-Malecki, B. (1990). "Colour constancy" in Mondrian patterns: A partial cancellation of physical chromaticity shifts by simultaneous contrast. *Vision Research* 30, 371–380.

Van De Weijer, J., and Gevers, T. (2005). Color constancy based on the grey-edge hypothesis., in *IEEE International Conference on Image Processing 2005*, (IEEE), II–722.

Van De Weijer, J., Gevers, T., and Gijsenij, A. (2007). Edge-based color constancy. *IEEE Transactions on image processing* 16, 2207–2214.

Wallach, H. (1948). Brightness constancy and the nature of achromatic colors. *Journal of experimental psychology* 38, 310.

Wang, D., Devaraj, J., and Wang, D. (2023). A Comprehensive Review of Deep Learning based Illumination Estimation. doi: 10.20944/preprints202302.0478.v1

Witzel, C., and Gegenfurtner, K. R. (2018). Color Perception: Objects, Constancy, and Categories. *Annual Review of Vision Science* 4, 475–499. doi: 10.1146/annurev-vision-091517-034231

Yu, H., Chen, K., Wang, K., Qian, Y., Zhang, Z., and Jia, K. (2020). Cascading convolutional color constancy., in *Proceedings of the AAAI Conference on Artificial Intelligence*, 12725–12732. Available at: https://aaai.org/ojs/index.php/AAAI/article/view/6966 (Accessed February 9, 2026).